\documentclass[final]{cvpr}

\usepackage{times}
\usepackage{epsfig}
\usepackage{graphicx}
\usepackage{amsmath}
\usepackage{amssymb}
\usepackage{pifont}
\usepackage{multirow}
\usepackage{bm}
\usepackage[dvipsnames]{xcolor}
\usepackage{colortbl} 
\usepackage{times,graphicx,amsmath,amssymb,booktabs,tabulary,multirow,overpic,xcolor}
\usepackage[caption=false]{subfig}
\definecolor{citecolor}{RGB}{34,139,34}

\usepackage[pagebackref=true,breaklinks=true,colorlinks,bookmarks=false]{hyperref}

\newcommand{\bd}[1]{\textbf{#1}}
\newcommand{\app}{\raise.17ex\hbox{$\scriptstyle\sim$}}

\newcolumntype{x}[1]{>{\centering\arraybackslash}p{#1pt}}

\newlength\savewidth\newcommand\shline{\noalign{\global\savewidth\arrayrulewidth
  \global\arrayrulewidth 1pt}\hline\noalign{\global\arrayrulewidth\savewidth}}
\newcommand{\tablestyle}[2]{\setlength{\tabcolsep}{#1}\renewcommand{\arraystretch}{#2}\centering\footnotesize}

\makeatletter
\newcommand{\printfnsymbol}[1]{%
  \textsuperscript{\@fnsymbol{#1}}%
}
\makeatother

\usepackage{bbm}
\definecolor{mygray}{gray}{0.6}
\definecolor{mygray-bg}{gray}{0.9}
\newcommand{\xmark}{\ding{55}}%
\newcommand{\cmark}{\ding{51}}%

\def\cvprPaperID{1453} 
\def\confYear{CVPR 2021}

\begin{document}

\title{Reformulating HOI Detection as Adaptive Set Prediction}

\author{Mingfei Chen\textsuperscript{\rm 1,3}\thanks{Equal contribution} \quad Yue Liao\textsuperscript{\rm 2}\printfnsymbol{1} \quad Si Liu\textsuperscript{\rm 2}\thanks{Corresponding author (liusi@buaa.edu.cn)} \quad Zhiyuan Chen\textsuperscript{\rm 3} \quad Fei Wang\textsuperscript{\rm 3}\quad Chen Qian\textsuperscript{\rm 3}\\
\large\textsuperscript{\rm 1} Huazhong University of Science and Technology \\ \textsuperscript{\rm 2} Institute of Artificial Intelligence, Beihang University \quad\textsuperscript{\rm 3} SenseTime Research}

\maketitle
\thispagestyle{empty}
\pagestyle{empty}
\begin{abstract}
Determining which image regions to concentrate is critical for Human-Object Interaction (HOI) detection. Conventional HOI detectors focus on either detected human and object pairs or pre-defined interaction locations, which limits learning of the effective features. In this paper, we reformulate HOI detection as an adaptive set prediction problem, with this novel formulation, we propose an Adaptive Set-based one-stage framework (AS-Net) with parallel instance and interaction branches. To attain this, we map a trainable interaction query set to an interaction prediction set with transformer. Each query adaptively aggregates the interaction-relevant features from global contexts through multi-head co-attention. Besides, the training process is supervised adaptively by matching each ground-truth with the interaction prediction. Furthermore, we design an effective instance-aware attention module to introduce instructive features from the instance branch into the interaction branch. Our method outperforms previous state-of-the-art methods without any extra human pose and language features on three challenging HOI detection datasets. Especially, we achieve over $31\%$ relative improvement on a large scale HICO-DET dataset. Code is available at \url{https://github.com/yoyomimi/AS-Net}.
\end{abstract}

\vspace{-2mm}\section{Introduction}
\vspace{-1mm}
Human-Object Interaction~(HOI) detection aims to identify HOI triplets $<$human, verb, object$>$ from a given image, it is an important step toward the high-level semantic understanding~\cite{gao2020interactgan,Liao_2020_CVPR,yu2020cross,hui2020linguistic,Huang_2020_CVPR,Jiang_2020_CVPR,gao2020adversarialnas,gao2020recapture,you2020greedynas}. Conventional HOI methods can be divided into two-stage methods~\cite{shen2018scaling,chao2018learning,gao2018ican,li2018transferable,Li_2019_CVPR,Gupta_2019_ICCV,Gao-ECCV-DRG,ren2020scene} and one-stage methods~\cite{Kim2020_unidet,liao2020ppdm}. Most two-stage methods detect instances~(humans and objects), and match the detected humans and objects one by one to form pair-wise proposals in the first stage. Next, in the second stage, such methods infer the interactions based on the features of cropped human-object pair-wise proposals. Two-stage methods have made great progress in HOI detection, however, their efficiency and effectiveness are limited by their serial architectures. With the development of one-stage object detectors, one-stage HOI detectors~\cite{Kim2020_unidet,liao2020ppdm} have raised a new fashion. Existing one-stage HOI detectors formulate HOI detection as a parallel detection problem, which detects the HOI triplets from an image directly. One-stage methods have delivered great improvements in both efficiency and effectiveness. 

\begin{figure}[t]
  \begin{center}
  \includegraphics[width=1.0\linewidth]{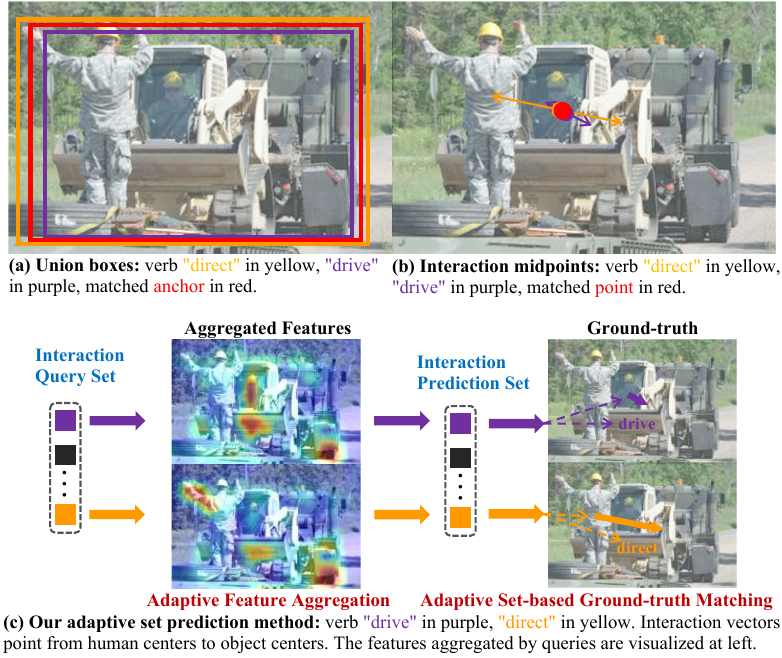}
  \end{center}\vspace{-4mm}
  \caption{Both the anchor-based~(a) and point-based~(b) one-stage methods infer two different interactions ``drive" and ``direct" are at similar location and concentrate on the similar features. Our set prediction method~(c) maps an interaction query set to an interaction prediction set by an interaction decoder. Then, interaction predictions are adaptively matched with ground-truth.
  To attain this, we first train a set of learnable embeddings as an interaction query set. Next, each interaction query adaptively aggregates the interaction-relevant features by co-attention. 
  Finally, we match each ground-truth with prediction for adaptive supervision. This mechanism empowers our method to accurately predict two interactions for ``drive" and ``direct". Best viewed in color.}\vspace{5mm}
  \label{fig:intro}
\end{figure}
Determining which regions to concentrate on is critical and challenging for HOI detectors.
To obtain essential features for interaction prediction, conventional two-stage methods usually involve extra features, \eg, human pose~\cite{shen2018scaling, feng2019turbo, Li_2019_CVPR, Gupta_2019_ICCV} and language~\cite{Xu_2019_CVPR, Gao-ECCV-DRG, Liu20a, kim2020detecting}. However, even with extra features, two-stage methods still focus on the detected instances that might be inaccurate, which are less adaptive and limited by the detected instances. One-stage methods partially alleviate these issues by inferring interactions directly from the whole image. Such methods intuitively define a location-relative medium to predict interactions, and can be mainly divided into anchor-based methods and point-based methods. Anchor-based methods~\cite{Kim2020_unidet} predict the interactions based on the union box of each pair-wise human and object instances. While point-based methods~\cite{liao2020ppdm} infer the interaction midpoint of each corresponding human-object pair. However, we argue that it is sub-optimal to predict the interaction through a pre-defined interaction location. 
Figure~\ref{fig:intro} illustrates an example. The interaction ``direct"~(in yellow) and ``drive"~(in purple) are quite different and thus require different visual features for interaction prediction. However, their union boxes are considerably overlapped~(Figure \ref{fig:intro} (a)), and their interaction midpoints are very close~(Figure \ref{fig:intro} (b)). Therefore, these one-stage methods concentrate on similar visual features for the two different interactions. 

To further address the limitation of interaction location in one-stage methods, we reformulate interaction detection as a set-based prediction problem. We define an interaction query set with several learnable embeddings, and an interaction prediction set. Each embedding in the query set is mapped by a transformer based interaction decoder to an interaction prediction set. By feeding the interaction query set into a multi-head co-attention module, we are able to adaptively aggregate features from global contexts. Our proposed method matches each ground-truth with the resembling interaction prediction for adaptive supervision. Therefore, our proposed method adaptively concentrates on the most suitable features for each prediction, free from the location limitation of conventional one-stage methods. As demonstrated in Figure~\ref{fig:intro} (c), our method aggregates arm features of the left person and pose features of the right person to make two different interaction predictions. The predictions are then matched with the ground-truth interaction ``direct" and ``drive" respectively.

To this end, we propose a novel Adaptive Set-based one-stage framework, namely AS-Net. Our AS-Net consists of two parallel branches: an instance branch and an interaction branch. Both branches leverage a transformer encoder-decoder structure, which utilize global features to perform set predictions. The instance branch predicts location and category for each instance, while the interaction branch predicts interaction vectors and their corresponding categories. The interaction vectors point from the centers of the human instances to the centers of the object instances. We obtain the predicted interaction triplets by matching each interaction vector from the interaction branch with the detected instances from the instance branch. Besides, we exploit an instance-aware attention module in a co-attention manner to perform branch aggregation. Specifically, this module aggregates information in the instance branch and introduces the aggregated features into the interaction branch. We also utilize semantic embeddings to perform more accurate human-object matching.

We test our proposed AS-Net on three datasets, \ie, HICO-Det~\cite{qi2018learning}, V-COCO~\cite{gupta2015visual}, and HOI-A~\cite{liao2020ppdm}. Our proposed AS-Net outperforms all the other algorithms among all datasets. In specific, our proposed AS-Net has gained $31\%$ relative improvement comparing to the previous state-of-the-art one-stage method~\cite{liao2020ppdm} on HICO-DET.

Our contributions can be concluded in the following three aspects:
\begin{itemize}
\item We formulate HOI detection as a set prediction problem, which breaks the instance-centric limitation and location limitation of the existing methods. Thereby, our method can adaptively concentrate on the most suitable features to improve the predicting accuracy.
\item We propose a novel one-stage transformer-based HOI detection framework, namely AS-Net. We also design an instance-aware attention module to introduce the information in the instance branch into the interaction branch. 
\item Without introducing any extra features, our method outperforms all the previous state-of-the-art methods, achieving $31\%$ relative improvement over the second best one-stage method on the HICO-DET dataset.
\end{itemize}

\vspace{-2mm}\section{Related Work}
\vspace{-1mm}
\noindent\textbf{Two-stage Methods.} 
Most conventional HOI detectors are in a two-stage manner. In the first stage, an object detector~\cite{girshick2015fast, ren2015faster, Dong_2020_CVPR} is applied to detect the instances. In the second stage, the cropped instance features are classified to obtain the interaction categories. In addition to the cropped instance features, previous methods leverage combined spatial features~\cite{chao2018learning, gkioxari2018detecting, gao2018ican, Gupta_2019_ICCV, Gao-ECCV-DRG, hou2020visual, zhong2020polysemy}, union box features~\cite{qi2018learning, Wan_2019_ICCV}, or context features~\cite{gao2018ican, Wang_2019_ICCV, Liu20a} to improve the accuracy of HOI detection. In order to concentrate on more interaction-relevant features, some methods utilize extra features, such as human pose~\cite{shen2018scaling, feng2019turbo, Li_2019_CVPR, Gupta_2019_ICCV}, human parts~\cite{Zhou_2019_ICCV, Wan_2019_ICCV, li2020detailed} and language features~\cite{Xu_2019_CVPR, Gao-ECCV-DRG, Liu20a, kim2020detecting}. However, the serial architectures of such two-stage methods impair the efficiency of HOI detection. Moreover, the prediction accuracy is usually limited by the results of instance detection.

\begin{figure*}[htb]
  \centering
  \includegraphics[width=0.99\linewidth]{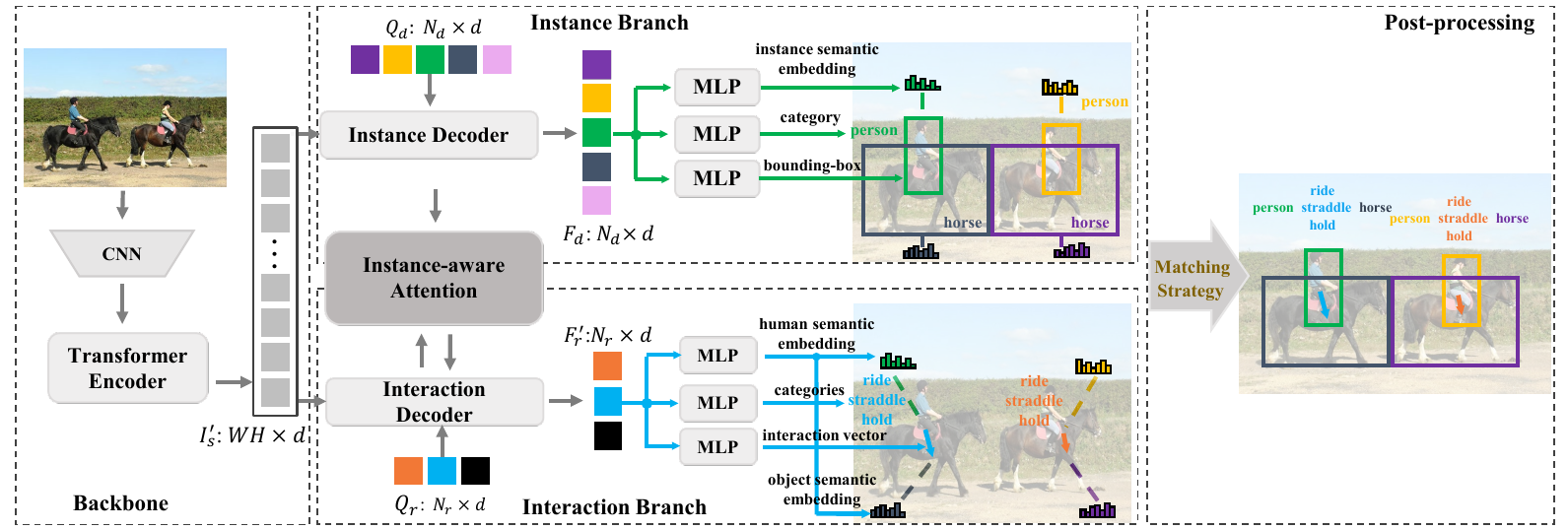}
  \vspace{-0.5mm}
  \caption{Overview of the proposed framework. First, a CNN and a transformer encoder are applied to extract the feature sequence with global contexts. Then two branches are built on the transformer decoder layers: a)~the instance branch transforms a set of learnable instance queries to an instance prediction set one by one b)~the interaction branch utilizes an interaction query set to estimate an interaction prediction set. The instance-aware attention module is designed to introduce the interaction-relevant instance features from the instance branch to the interaction branch. At the end, the detected instances are matched with the interaction predictions to infer the HOI triplets.}
  \label{framework}\vspace{-2mm}
\end{figure*}

\noindent\textbf{One-stage Methods.} 
Recently, one-stage HOI detection methods with higher efficiency~\cite{Kim2020_unidet, liao2020ppdm} have attracted increasing attention. Most one-stage methods extract features with a bottom-up structure\cite{newell2016stacked, Yu_2018_CVPR}, and detect the HOI triplets in parallel from an image directly. Specifically, the one-stage methods can be divided into anchor-based methods~\cite{Kim2020_unidet} and point-based methods~\cite{liao2020ppdm} according to the manners of their interaction prediction. The anchor-based methods predict the interactions based on each union box. The point-based methods perform inference at each interaction key point, such as the midpoint of each corresponding human-object pair. Though breaking the limitation from instance detection, such methods which pre-assign each ground-truth interaction to the predictions, are still non-adaptive and limited by the interaction locations.
\vspace{-2.5mm}\section{Methods}
\vspace{-1.5mm}
HOI detection aims to predict the triplet of $<$human, verb, object$>$, which contains a pair of bounding-boxes for a human and an object, and a corresponding verb category. In this paper, we reformulate HOI detection as a set prediction problem, and propose an Adaptive Set-based one-stage Network (AS-Net). 

Our AS-Net builds on a transformer encoder-decoder architecture and makes parallel set-based predictions for the HOI triplets. As illustrated in Figure~\ref{framework}, our proposed AS-Net consists of four parts. We first utilize a backbone~(Section \ref{sec:backbone}) to extract the visual feature sequence with global contexts. The instance~(Section \ref{sec:instance}) and interaction branches~(Section \ref{sec:interaction}) following the backbone parallelly detect an instance and interaction prediction set from the feature sequence respectively. In order to intensify the instance features that are valuable for interaction inference, we design an instance-aware attention module~(Section \ref{sec:IA}) to perform branch aggregation. Specifically, we introduce semantic embeddings (Section~\ref{sec:se}) in instance and interaction branches for more accurate triplet prediction. At the end, we match detected instances and interactions to obtain the final HOI triplets~(Section \ref{sec:training and postprocess}).
\vspace{-1.5mm}\subsection{Backbone} \label{sec:backbone}\vspace{-1mm}
We define the backbone by combining a CNN and a transformer encoder to extract the image features. The encoder is in a multi-layer manner, where each layer comprises a multi-head self-attention module and a two-layer Feed-Forward Network~(FFN). For a given image, we first extract a visual feature map $\bm{I} \in \mathbb{R}^{W\times H\times C}$ using the CNN. Then we utilize a $1 \times 1$ convolution to reduce the channel dimension of the visual feature map from $C$ to $d$, and reshape such feature map as a feature sequence $\bm{I}_s \in \mathbb{R}^{WH\times d}$. Next, we feed the feature sequence to the encoder which refines the feature sequence by introducing global contexts into the output feature sequence $\bm{I}_s' \in \mathbb{R}^{WH\times d}$.
\vspace{-1.5mm}\subsection{Instance Branch} \label{sec:instance}\vspace{-1mm}
The instance branch is designed to localize and classify the instances. Following the detector DETR~\cite{carion2020endtoend}, our instance branch consists of a multi-layer transformer decoder and several FFN heads. Each layer of the decoder is comprised of a self-attention module, and a multi-head co-attention module. 
The input of each decoder layer is the summation of a learnable positional embedding sequence $\bm{Q}_d \in \mathbb{R}^{N_d\times d}$ and the output of last layer. Except for the first layer where there is no output of last layer, we added zeros to the learnable positional embedding sequence.
We first feed the input into the self-attention module. Then the multi-head co-attention module adaptively aggregates the key contents from $\bm{I}_s'$ to $\bm{F}_d \in \mathbb{R}^{N_d\times d}$, where we take $\bm{Q}_d$ with the output of self-attention as queries, and $\bm{I}_s'$ with the corresponding fixed positional encodings~\cite{pmlr-v80-parmar18a} as keys. 
There is an FFN head on top of each decoder layer to decode a set of instance predictions from $\bm{F}_d$. The FFN head comprises three independent sub-branches. One to predict the normalized bounding-box in $(cx, cy, w, h)$ format for each detected instance. Another to infer a $(L_d+1)$-dimensional scores for $L_d$ categories, where the last dimension refers to the no-object~($\varnothing$) category. The other to generate a distinctive semantic embedding $\bm\varepsilon \in \mathbb{R}^K$ for each instance, which will be explained in Section \ref{sec:se}. Each sub-branch constitutes of one or several perception layers. The FFN head of each decoder layer shares the same weights.

\vspace{0.5mm}\noindent\textbf{Training.} For the set-based training process, we first find a one-to-one bipartite matching between the detected instance set $\hat y$ and the ground-truth $y$~(padded with no-instance $\varnothing$ to a set of size $N_d$). To this end, we deploy a matching loss, which is the summation of bounding-box loss and category semantic distance between instance and all ground-truth bounding-boxes. Following~\cite{carion2020endtoend}, the bounding-box loss is composed of a $l_1$ loss and a GIoU loss~\cite{rezatofighi2019generalized}. The category semantic distance is the negative of summation of the predicted scores for each ground-truth category.

The universal index permutation set of $N_d$ predictions is denoted as ${S}_{N_d}$. We consider $\hat \sigma_d \in {S}_{N_d}$ that minimizes the summation of all the matching cost $\mathcal{L}_{\operatorname{match}}(\hat {y}_{\sigma_d(i)}, y_i)$ as the optimal index permutation of the detected instance set, which we adopt the Hungarian algorithm~\cite{hungarian} to calculate. The $i$-th element of the index permutation $\sigma_d \in {S}_{N_d}$ is defined as $\sigma_d(i)$, and the $\hat \sigma_d$ is formulated as:
\vspace{-1mm}\begin{equation}
\small
\hat\sigma_d=\underset{\sigma_d \in {S}_{N_d}}{\arg \min } \sum_{i=1}^{N_d} \mathbbm{1}_{y_i \neq \varnothing} \mathcal{L}_{\operatorname{match}}(\hat {y}_{\sigma_d(i)}, y_i).
\label{eq:sigma}
\end{equation}

For the instance prediction with index permutation $\hat{\sigma}_d(i)$, the predicted bounding-box and category are represented as $\hat{b}_{\hat{\sigma}_d(i)}$ and $\hat{p}_{\hat{\sigma}_d(i)}$ respectively. We follow the DETR detector~\cite{carion2020endtoend} to construct the set-based instance detection loss $\mathcal{L}_{\operatorname {ins}}$:
\begin{equation}
\small
\vspace{-2mm}
\hspace{-1.5mm}\mathcal{L}_{\operatorname {ins}}=\sum_{i=1}^{N_d}[-\log \hat{p}_{\hat{\sigma}_d(i)}(c_{i})+\mathbbm{1}_{\{c_{i} \neq \varnothing\}} \mathcal{L}_{\operatorname {box }}(b_{i}, \hat{b}_{\hat{\sigma}_d(i)})],
\end{equation}
where $b_i$ and $c_i$ denotes the bounding-box and category of the matched ground-truth instance respectively, $\hat{p}_{\hat{\sigma}_d(i)}(c_{i})$ is the confidence score for category $c_i$.
\vspace{-1.5mm}\subsection{Interaction Branch}\vspace{-1mm} \label{sec:interaction}
The interaction branch predicts the interaction vectors and categories for each interaction. Its architecture is similar to the instance branch, which constitutes a multi-layer transformer decoder and several FFN heads. Each decoder layer utilizes several interaction query set~$\bm{Q}_r$ to aggregate the corresponding key contents $\bm{F}_r \in \mathbb{R}^{N_r\times d}$ from the shared feature sequence $\bm{I}_s'$. 
Each decoder layer is equipped with a FFN head as the instance branch. Each FFN head is also split into three sub-branches. For each interaction prediction, we predict a $4$-dimensional interaction vector with categories, and two semantic embeddings, \ie, $\bm\varepsilon^h \in \mathbb{R}^K$ and $\bm\varepsilon^o \in \mathbb{R}^K$ for the corresponding human and object instances respectively. The interaction vector points from the normalized human center $(x^{h}_{ct}, y^{h}_{ct})$ to the object center $(x^{o}_{ct}, y^{o}_{ct})$. Considering there might exist multiple interactions for the same human-object pair, we use a multi-label classifier to predict a score for each verb category respectively.

\vspace{0.5mm}\noindent\textbf{Training.} We denote the ground-truth interaction as $t = (v, z)$, where $v$ is the interaction vector of $t$, and $z$ indicates the $L$ ground-truth interaction categories of $t$. We compute the matching loss between $t$ and each predicted interaction $\hat {t} = (\hat v, \hat{z})$, where $\hat v$ refers to the predicted interaction vector and $\hat{z}$ indicates the confidence scores of the interaction categories. The matching cost $\mathcal{L}_{\operatorname{match}}(\hat {t}_{\sigma_r(i)}, t_i)$ can be computed by:
 \vspace{-1mm}\begin{equation}
	\mathcal{L}_{\operatorname{match}}(\hat {t}_{\sigma_r(i)}, t_i)= \|v_{i}-\hat v_{\sigma_r(i)} \|_{1} + \sum_{l=1}^{L}-\frac{1}{1+e^{-\hat {z}_{\sigma_r(i)}(z_l)}},
\end{equation}
where $\hat {z}_{\sigma_r(i)}(z_l)$ refers to the score for the $l$-th ground-truth interaction category $z_l$ of $t$. Similar to the set-based training process for the instance branch, we utilize the Hungarian algorithm~\cite{hungarian} to find the optimal index assignment $\hat \sigma_r$ for the predicted interaction set \emph{w.r.t.} the ground-truth.

For the interaction prediction with index $\hat{\sigma}_r(i)$, we define the predicted interaction vector and categories as $\hat{v}_{\hat\sigma_r(i)}$ and $\hat {z}_{\hat{\sigma_r}(i)}$ respectively, and the matched target interaction vector and categories are $v_i$ and ${z}_i$ respectively. To balance the ratio between the positive and negative samples for each classifier, we apply Focal loss~\cite{lin2017focal}, denoted as $\mathcal{L}_{\operatorname{cls}}$, for the training of interaction classification. Besides, we adopt $l_1$ loss, denoted as $\mathcal{L}_{\operatorname{reg}}$, for the regression of interaction vectors. The interaction loss $\mathcal{L}_{\operatorname{int}}$ is calculated as:
\begin{equation}
\begin{aligned}
	\mathcal{L}_{\operatorname{int}}&=\sum_{i=1}^{N_r} ~[\lambda_{\operatorname{cls}}\mathcal{L}_{\operatorname{cls}}({z}_i, \hat {z}_{\hat{\sigma}_r(i)}) \\ &+ \mathbbm{1}_{\{{z}_{i} \neq \varnothing\}} \lambda_{\operatorname{reg}}\mathcal{L}_{\operatorname{reg}}(v_i, \hat{v}_{\hat\sigma_r(i)})],	
    \end{aligned}
	\end{equation}
where $\lambda_{\operatorname{cls}}$ and $\lambda_{\operatorname{reg}}$ are the weight coefficients of $\mathcal{L}_{\operatorname{cls}}$ and $\mathcal{L}_{\operatorname{reg}}$ respectively.

\vspace{0.5mm}\noindent\textbf{Analysis.}
Adaptation is involved in the interaction prediction from two aspects.
First, for each interaction query, we apply multi-head co-attention to aggregate information from each element in the feature sequence. Hence, each query can adaptively aggregate the interaction-relevant visual features. 
Second, instead of pre-assigning each ground-truth to the corresponding prediction, we consider both the predicted interaction vectors and categories to match each ground-truth interaction with the resembling prediction. Therefore, each interaction prediction can be supervised by the most suitable ground-truth more adaptively.

\vspace{-1.5mm}\subsection{Instance-aware Attention} \label{sec:IA}\vspace{-1mm}
We construct an instance-aware attention module between each instance and interaction layer to emphasize relevant instance features for interaction prediction. 

First, we compute an affinity score map $\bm{A} \in \mathbb{R}^{(N_r \times N_d)}$ between the instance features $\bm{F}_d$ and the interaction features $\bm{F}_r$:
\begin{equation}
\bm{A}=\frac {(\bm{W}_r \bm{F}_r+b_r)({\bm{W}_d \bm{F}_d+b_d})^{\top}}{\sqrt{d}}.
\end{equation}
We then apply Softmax to obtain the instance-aware attention weight matrix $\bm{M} \in[0,1]^{(N_r \times N_d)}$:
\begin{equation}
\bm{M}_{ij} = \frac{exp(\bm{A}_{ij})}{\sum_{j=1}^{N_d}exp(\bm{A}_{ij})},
\end{equation}where $\bm{M}_{ij}$ refers to the attention weight of the $j$-th detected instance with respect to the $i$-th predicted interaction. The final output interaction features $\bm{F}_{r}^\prime \in \mathbb{R}^{(N_r \times d)}$ of the instance-aware attention module is formulated as:
\begin{equation}
\bm{F}_{r}^\prime = \bm{M}(\bm{W}_d^\prime \bm{F}_d+b_d^\prime) + \bm{F}_r.
\end{equation}

\vspace{-1.5mm}\subsection{Semantic Embedding} \label{sec:se}\vspace{-1mm}
The interaction vectors are not pointing to a instance directly, instead, they point to a region. Instead of matching which only employs location indication from the interaction vectors, we introduce semantic embeddings inferred by an MLP block in our matching strategy. We infer semantic embeddings $\bm\varepsilon$ from $\bm{F}_d$ for each detected instance in instance branch. And in the interaction branch, two semantic embeddings $\bm\varepsilon^h$ and $\bm\varepsilon^o$ are inferred from $\bm{F}_r'$, one for the human instance and another for the object instance for each prediction. 

In the training process, the semantic embeddings of different instances are pushed away from each other. The push procedure can be described as:
\begin{equation}
\small
\vspace{-2mm}
	\mathcal{L}_{\operatorname{push}}=\sum_{i=1}^{ | \hat\sigma_d  |-1} \sum_{j=i+1}^{ | \hat\sigma_d  |} [\max (0, t-\|\bm \varepsilon_{\hat \sigma_d(i)}-\bm \varepsilon_{\hat \sigma_d(j)}\| )]^2,
\end{equation}
where $ | \hat \sigma_d  |$ refers to the total number of the ground-truth instances, and $\bm\varepsilon_{\hat \sigma_d(i)}$ refers to the semantic embedding of the predicted instance matched to the $i$-th target instance. If the $l_2$ distance between two semantic embeddings are more than a threshold $t$, we consider two embeddings are separate enough and set $\mathcal{L}_{\operatorname{push}}$ to $0$.

We pull the semantic embeddings that refer to the same instance towards each other:
 \begin{equation}
    \small
    \vspace{-2mm}
	\hspace{-1.5mm}\mathcal{L}_{\operatorname{pull}} = \sum_{i=1}^{| \hat\sigma_r |}( \|\bm \varepsilon^{h}_{\hat \sigma_r(i)}-\bm \varepsilon_{\hat\sigma_d(h_i)}\|^{2} + \|\bm \varepsilon^{o}_{\hat \sigma_r(i)}-\bm \varepsilon_{\hat\sigma_d(o_i)}\|^{2}),
\end{equation}
where we denote the predicted human semantic embedding as $\bm \varepsilon_{\hat\sigma_r(i)}^h$ and the object embedding as $\bm \varepsilon_{\hat\sigma_r(i)}^o$ for the interaction prediction with index $\hat\sigma_r(i)$. The semantic embedding $\bm \varepsilon_{\hat\sigma_d(h_i)}$ and $\bm \varepsilon_{\hat\sigma_r(i)}^h$ refer to the same human instance in the instance and interaction branches respectively. Similarly, $\bm \varepsilon_{\hat\sigma_d(o_i)}$ and $\bm \varepsilon_{\hat\sigma_r(i)}^o$ refer to the same object instance. $ | \hat \sigma_r  |$ refers to the total number of the ground-truth interactions.

\vspace{-1.5mm}\subsection{Training Loss and Post-processing} \label{sec:training and postprocess}\vspace{-1mm}
The target loss is the weighted sum of the losses mentioned above:
\vspace{-1mm}
 \begin{equation}
 \small
	\mathcal{L} = \mathcal{L}_{\operatorname{ins}} + \mathcal{L}_{\operatorname{int}} + \lambda_{\operatorname{emb}} (\mathcal{L}_{\operatorname{pull}} + \mathcal{L}_{\operatorname{push}} ),
\vspace{-2mm}
\end{equation}
where $\lambda_{\operatorname{emb}}$ is a hyper-parameter to balance different loss.
\begin{table*}[htb!]
  \begin{center}
  \small
  \resizebox{1.0\textwidth}{!}{%
  \begin{tabular}{ccccc|ccc|ccc}
    \hline
    &&Finetune&&&\multicolumn{3}{c|}{Default} & \multicolumn{3}{c}{Know Object} \\
  Method    &Backbone  &Detection    &Extra   &Time (ms)~/~FPS   & Full & Rare & Non-Rare & Full  & Rare & Non-Rare\\
  \hline
  Two-stage Method:        &      &     &       &       & & & &  & &\\
  InteractNet~\cite{gkioxari2018detecting}	&ResNet-50-FPN	&\xmark&\xmark		& 145~/~6.90		&9.94	&7.16	&10.77       &-	&-	&-\\
  GPNN~\cite{qi2018learning}	&Res-DCN-152		&\xmark&\xmark		& -				&13.11	&9.34	&14.23       &-	&-	&-\\
  iCAN~\cite{gao2018ican}	        &ResNet-50		&\xmark&\xmark		& 204~/~4.90		&14.84	&10.45	&16.15       &16.26	&11.33	&17.73\\
  No-Frills~\cite{Gupta_2019_ICCV}		&ResNet-152		&\xmark&\emph{P}	& 494~/~2.02		&17.18	&12.17	&18.68       &-	&-	&-\\ 
  PMFNet~\cite{Wan_2019_ICCV}		&ResNet-50-FPN	&\xmark&\emph{P}	& 253~/~3.95		&17.46	&15.65	&18.00       &20.34	&17.47	&21.20\\ 
  DRG~\cite{Gao-ECCV-DRG}			&ResNet-50-FPN	&\xmark&\emph{L}	& 200~/~5.00				&19.26	&17.74	&19.71		 &23.40	&21.75	&23.89\\
  IP-Net~\cite{wang2020learning}		&Hourglass-104	&\xmark&\xmark		& -			&19.56	&12.79	&21.58      	 &22.05	&15.77	&23.92\\
  VSGNet~\cite{Ulutan_2020_CVPR}		&ResNet-152		&\xmark&\xmark		& 312~/~3.21				&19.80	&16.05	&20.91       &-	&-	&-\\ 
  PD-Net~\cite{zhong2020polysemy}		&ResNet-152-FPN	&\xmark&\emph{L}	& -				&20.81	&15.90	&22.28       &24.78	&18.88	&26.54\\ 
  DJ-RN~\cite{li2020detailed}			&ResNet-50		&\xmark&\emph{P}	& -				&21.34	&18.53	&22.18       &23.69	&20.64	&24.60\\
  \hline 
  One-stage Method:        &           &       &       & & & &  & &\\
  UnionDet~\cite{Kim2020_unidet}			&ResNet-50-FPN	&\cmark&\xmark		& 78~/~12.82		&17.58	&11.72	&19.33       &19.76	&14.68	&21.27\\
  PPDM-Hourglass~\cite{liao2020ppdm}		&Hourglass-104	&\cmark&\xmark		& 71~/~14.08	&21.94	&13.97	&24.32       &24.81	&17.09	&27.12\\
  \cellcolor{mygray-bg}AS-Net*		&\cellcolor{mygray-bg}ResNet-50		&\cellcolor{mygray-bg}\xmark&\cellcolor{mygray-bg}\xmark		&\cellcolor{mygray-bg}71~/~14.08	&\cellcolor{mygray-bg}24.40	&\cellcolor{mygray-bg}22.39	&\cellcolor{mygray-bg}25.01       &\cellcolor{mygray-bg}27.41 &\cellcolor{mygray-bg}25.44  &\cellcolor{mygray-bg}28.00\\
  \cellcolor{mygray-bg}AS-Net~~				&\cellcolor{mygray-bg}ResNet-50		&\cellcolor{mygray-bg}\cmark&\cellcolor{mygray-bg}\xmark		&\cellcolor{mygray-bg}\textbf{71~/~14.08}			&\cellcolor{mygray-bg}\textbf{28.87}	&\cellcolor{mygray-bg}\textbf{24.25} 	&\cellcolor{mygray-bg}\textbf{30.25}       &\cellcolor{mygray-bg}\textbf{31.74}	&\cellcolor{mygray-bg}\textbf{27.07}	&\cellcolor{mygray-bg}\textbf{33.14}\\
  \hline          
  \end{tabular}}
  \end{center}
   \vspace{-2mm}
    \caption{\textbf{Performance comparison on the HICO-DET test set.} The `P', `L' represent human pose information and the language feature, respectively. * denotes freezing the instance detection related parameters pretrained on the MS-COCO dataset. Our one-stage model with a high inference speed of $71$~ms~/~$14.08$~FPS outperforms all previous work by a large margin.}
  \label{tb:hico}\vspace{-3mm}
  \end{table*}

During the post-processing, we first match the detected human instances with object instances based on our predicted interaction vectors and semantic embeddings. A good human-object interaction match should meet the following three requirements: 1) the normalized center of the matched human/object instances is close to the start and the end point of the interaction vector respectively; 2) the matched instances have high confidence scores on their predicted categories; 3) the semantic embedding referring to the same matched instances are similar to each other.

We consider all detected instances as object instances. For each predicted interaction vector $\hat v = (\hat x^h_{ct}, \hat y^h_{ct}, \hat x^o_{ct}, \hat y^o_{ct})$, the matching distance $\mathcal D$ can be calculated as:
\begin{equation}
\small
\begin{split}
\mathcal{D} = &(|\tilde{x}^h_{ct}-\hat x^h_{ct}| + 1)(|\tilde{y}^h_{ct}-\hat y^h_{ct}| + 1) \\ &
(|\tilde{x}^{o}_{ct}-\hat x^o_{ct}| + 1)(|\tilde{y}^o_{ct}-\hat y^o_{ct}| + 1),
\end{split}
\label{eq:match_offset}
\end{equation}
where $(\tilde{x}^h_{ct}, \tilde{y}^h_{ct})$, $(\tilde{x}^o_{ct}, \tilde{y}^o_{ct})$ refers to the center of the detected human and object instances, with a confidence score of $s^h$ and $s^o$, respectively.

When the semantic embeddings are introduced for matching, given the human~($\bm \varepsilon_{h}$) and object~($\bm \varepsilon_{o}$) semantic embedding of the instance branch, the embedding matching distance $\mathcal R$ for the predicted human~($\bm {\hat \varepsilon}_h$) and object~($\bm {\hat \varepsilon}_o$) semantic embedding of the interaction branch can be defined as: 
\begin{equation}
\small
\mathcal{R} = (\|\bm \varepsilon_{h} - \bm {\hat \varepsilon}_h  \| + 1 ) (\|\bm \varepsilon_{o} - \bm {\hat \varepsilon}_o  \| + 1 ).
\label{eq:match_emb}
\end{equation}

The final matching cost is calculated as $\frac{\mathcal{D}\mathcal{R}}{s^h s^o}$. We match the detected instances with the minimum matching cost to each interaction prediction. The HOI confidence score for each predicted triplet is the product of the interaction category score, and the matched instance scores $s^h$ and $s^o$. Triplets with top $N$ confidence scores are preserved as the final HOI triplet predictions.
\section{Experiments}
\vspace{-1.5mm}\subsection{Datasets and Metrics}\vspace{-1mm}
\noindent\textbf{Datasets.}
To verify the effectiveness of our model, we conduct experiments on three HOI detection datasets HICO-DET~\cite{chao2018learning}, V-COCO~\cite{gupta2015visual} and HOI-A~\cite{liao2020ppdm}. HICO-DET contains $38, 118$ images for training and $9, 658$ images for testing, contains the same $80$ object categories as MS-COCO~\cite{lin2014microsoft} and $117$ verb categories. The objects and verbs form $600$ classes of HOI triplets. V-COCO provides $2, 533$ images for training, $2, 867$ images for validating and $4, 946$ images for testing. V-COCO is derived from MS-COCO dataset, annotated with $29$ action categories. HOI-A dataset consists of $38, 668$ annotated images, $11$ kinds of objects and $10$ action categories.

\vspace{0.5mm}\noindent\textbf{Metrics.}
Following ~\cite{chao2018learning}, the mean average precision~(mAP) is adopted as evaluation metric. For one positive predicted HOI triplet $\langle$human, verb, object$\rangle$, both the predicted human and object bounding-boxes have IoUs greater than $0.5$ \emph{w.r.t.} the ground-truth boxes, with the correct predicted verb simultaneously.


\begin{table}[htb!]
  \begin{center}
  \small
  \begin{tabular}{cccc}
  \hline
  Method    &Backbone      &Extra  & $\operatorname{mAP}_{role}$\\
  \hline\hline
  Two-stage Method:     &               &                       & \\
  InteractNet~\cite{gkioxari2018detecting}	&ResNet-50-FPN	&\xmark				    &40.0\\
  GPNN \etal~\cite{qi2018learning}	&Res-DCN-152    &\xmark				    &44.0\\
  iCAN~\cite{gao2018ican}	        &ResNet-50		&\xmark				    &45.3\\
  DRG~\cite{Gao-ECCV-DRG}			&ResNet-50-FPN	&\emph{L}		        &51.0\\
  IP-Net~\cite{wang2020learning}		&Hourglass-104	&\xmark				    &51.0\\
  VSGNet~\cite{Ulutan_2020_CVPR}		&ResNet-152		&\xmark				    &51.8\\ 
  PMFNet~\cite{Wan_2019_ICCV}		&ResNet-50-FPN	&\emph{P}               &52.0\\ 
  PD-Net~\cite{zhong2020polysemy}		&ResNet-152-FPN	&\emph{L}               &52.6\\
  FCMNet~\cite{Liu20a}		&ResNet-50		&\xmark		            &53.1\\
  \hline 
  One-stage Method:     &               &                       &\\
  UnionDet~\cite{Kim2020_unidet}		&ResNet-50-FPN	&\xmark			        &47.5\\
  \cellcolor{mygray-bg}AS-Net*			&\cellcolor{mygray-bg}ResNet-50  	&\cellcolor{mygray-bg}\xmark			        &\cellcolor{mygray-bg}\textbf{53.9}\\
  \hline
  \end{tabular}
  \end{center}
  \vspace{-2mm}
    \caption{\textbf{Performance comparison on the V-COCO test set}. The `P', `L' represent the human pose information and the language feature, respectively. * denotes freezing the instance detection related parameters pretrained on the MS-COCO dataset.}
  \label{tb:vcoco}
\end{table}
\begin{table}[htb!]
  \begin{center}
  \small
  \begin{tabular}{cccc}
  \hline
  Method    &Backbone      &Extra  &mAP\\
  \hline\hline
  Two-stage Method:     &               &                       & \\
  iCAN~\cite{gao2018ican}	        &ResNet-50		&\xmark				    &44.23\\
  TIN~\cite{li2018transferable} &ResNet-50   &\emph{P}   & 48.64\\
  Faster Interaction Net~\cite{picleadboard}       &ResNet-50      &\xmark    &56.93\\
  GMVM~\cite{picleadboard}       &ResNet-50      &\emph{P}    &60.26\\
  C-HOI~\cite{zhou2020cascaded}       &ResNet-50      &\emph{P}   &66.04\\
  \hline 
  One-stage Method:     &               &                       &\\
  PPDM-DLA~\cite{liao2020ppdm}		&DLA-34	&\xmark	&67.03\\
  PPDM-Hourglass~\cite{liao2020ppdm}		&Hourglass-104	&\xmark	&71.23\\
  \cellcolor{mygray-bg}AS-Net			&\cellcolor{mygray-bg}ResNet-50  	&\cellcolor{mygray-bg}\xmark			        &\cellcolor{mygray-bg}\textbf{72.19}\\
  \hline
  \end{tabular}
  \end{center}
  \vspace{-2mm}
    \caption{\textbf{Performance comparison on the HOI-A test set}. The `P' represents the extra human pose or body parts information.}
  \label{tb:hoia}
\end{table}
\begin{table*}[t]
\centering
\subfloat[\textbf{Matching Strategy:} Analysis of different matching strategies, \ie, interaction vector and semantic embeddings.\label{tab:ablation:match}]{
\tablestyle{2.5pt}{1.05}\begin{tabular}{c|x{28}x{28}x{38}}
 \scriptsize \emph{Strategy} & Full & Rare & Non-Rare\\
\shline
 \scriptsize \emph{Vector} & 28.56 & 24.13 & 29.88\\
 \scriptsize \emph{Embedding} & 28.65 & 23.95 & 30.05\\\hline
 \scriptsize \emph{Combined} & \bd{28.87} & \bd{24.25} & \bd{30.25}\\
 \multicolumn{4}{c}{~}\\
\end{tabular}}\hspace{3mm}
\subfloat[\textbf{Dimension of Semantic Embeddings:} Choice of dimension of semantic embeddings.\label{tab:ablation:K}]{
\tablestyle{4.8pt}{1.05}\begin{tabular}{c|x{27}x{27}x{37}|x{40}}
\scriptsize \emph{$K$}  & Full & Rare & Non-Rare & \#Parameters \\
\shline
 \emph{4} & 28.21 & 22.65 & 29.87 &52.527~M\\
 \emph{8} & \bd{28.87} & \bd{24.25} & \bd{30.25} &52.530~M\\
 \emph{16} & 28.36 & 23.08 & 29.93 &52.537~M\\
 \emph{32} & 28.70 & 23.83& 30.16 &52.549~M\\
\end{tabular}}\hspace{3mm}
\subfloat[\textbf{Weight Coefficient $\mathbf{\lambda_{\operatorname{emb}}}$:} The effects of different settings of loss weight.\label{tab:weight}]{
\tablestyle{2.5pt}{1.05}\begin{tabular}{c|x{28}x{28}x{38}}
 \scriptsize \emph{$\lambda_{\operatorname{emb}}$} & Full & Rare & Non-Rare\\
\shline
 \scriptsize \emph{0.05} & 28.31 & 23.65 & 29.70\\
 \scriptsize \emph{0.1} & \bd{28.87} & \bd{24.25} & \bd{30.25}\\
 \scriptsize \emph{0.5} & 27.84 & 21.71 & 29.67\\
 \multicolumn{4}{c}{~}\\
\end{tabular}}\hspace{3mm}

\vspace{-3mm}
\subfloat[\textbf{Component Analysis:} Results of the variants with various components, \ie, interaction branches~(Int), instance-aware attention module~(IA Attn) and semantic embeddings~(emb). \label{tab:ablation:component}]{
\tablestyle{2.2pt}{1.05}\begin{tabular}{c|c|c|c|x{32}x{32}x{40}|x{40}}
 & \scriptsize Decoder Layers & \scriptsize Embeddings & \scriptsize IA Attention
 & Full & Rare & Non-Rare & \#Parameters \\
\shline
\emph{Single Branch} &$6\times$ &\xmark &- & 25.91 & 17.88 & 28.31 & 41.44~M\\
\hline
 \emph{Basic Model, Int$\times6$} &$6\times$ &\xmark &- & 27.52 & 22.04 &29.16 &50.94~M\\
 \emph{+ IA Attn$\times6$, Int w/o emb}$\times6$ & $6\times$ &\xmark  &$6\times$  & 27.96 & 23.01 &29.44 &52.13~M\\
\hline
\emph{+ Int w/ emb$\times6$} &$6\times$& \cmark &- & 27.75 &22.71 &29.25 & 51.34~M\\
\emph{+ IA Attn$\times3$, Int w/ emb$\times6$} &$6\times$ & \cmark & $3\times$ & 28.39 & 24.02 &29.70 &51.94~M\\
\emph{+ IA Attn$\times3$, Int w/ emb$\times3$} &$3\times$ & \cmark & $3\times$ & 28.63 & 23.61 &30.13 &47.20~M\\
\emph{+ IA Attn$\times6$, Int w/ emb$\times6$} &$6\times$ & \cmark & $6\times$ & \bd{28.87} & \bd{24.25} &\bd{30.25} &52.53~M\\
\end{tabular}}\\
\vspace{1mm}
\caption{Ablation studies of our proposed model on the HICO-DET test set.}
\label{tab:ablations}\vspace{-4mm}
\end{table*}
\vspace{-1.5mm}\subsection{Implementation Details}\vspace{-1mm}
Our implementation is based on two parallel $6$-layer transformer decoders with a shared backbone, where the backbone is built on ResNet-$50$~\cite{he2016deep} with a $6$-layer self-attention encoder. Following the detector DETR~\cite{carion2020endtoend}, we infer $N_d=100$ instances based on the aggregated interaction-relevant contents $\bm{F}_d \in \mathbb{R}^{100\times 256}$ on the instance branch. On the interaction branch, we infer a set of $N_r=16$ interaction vectors with categories from $\bm{F}_r' \in \mathbb{R}^{16\times 256}$. Moreover, the predicted semantic embeddings are all with a dimension of $K=8$. After the matching process~\ref{sec:training and postprocess}, the top $N=100$ predictions are finally preserved.

During training, we resize the shortest side of the input image to the range $[480, 800]$, and the longest side is no more than $1, 333$. We set the weight coefficients $\lambda_{\operatorname{cls}}$, $\lambda_{\operatorname{reg}}$ and $\lambda_{\operatorname{emb}}$ in Section~\ref{sec:training and postprocess} to $1$, $2$ and $0.1$ respectively. The model is trained with AdamW~\cite{loshchilov2018decoupled} for $90$ epochs on the HICO-DET and HOI-A dataset, and for $75$ epochs on the V-COCO dataset, with a learning rate of $10^{-4}$ decreased by $10$ times at the $70$th epoch.
All the instance detection related parameters~(backbone and the instance decoder layers) which are pretrained on the MS-COCO dataset, are frozen on the V-COCO dataset and trained with a learning rate of $10^{-5}$ on the other two datasets. Our experiments are all conducted on the GeForce GTX $1080$Ti GPU and CUDA $9.0$, with a batchsize of $64$ on $32$ GPUs. 

\vspace{-1.5mm}\subsection{Comparing to State-of-the-art}
\vspace{-1mm}
We conduct experiments on three HOI detection benchmarks to verify the effectiveness of our AS-Net. It is shown in Table~\ref{tb:hico}, Table~\ref{tb:vcoco} and Table~\ref{tb:hoia} that our AS-Net has achieved state-of-the-art across all the three benchmarks. Specifically, on the HICO-DET dataset, comparing to the previous state-of-the-art one-stage method PPDM~\cite{liao2020ppdm} which adopts Hourglass-104 as backbone, our AS-Net has achieved a $31\%$ performance gain with a relatively light-weight backbone, \emph{i.e.}, ResNet-$50$. Since the object detectors in two-stage methods are purely trained on MS-COCO, which does not fine-tune on HICO-DET, thus we also show the result when only training the interaction branch for fair comparison. In this setting, our AS-Net* has achieved $24.40\%$ mAP, which is superior to all existing two-stage methods, and has achieved above $3\%$ mAP improvements.

We compare our results on the V-COCO dataset with other state-of-the-art methods. Freezing the instance detection related parameters pretrained on the MS-COCO dataset, we only train the remaining parameters of our model. As shown in Table~\ref{tb:vcoco}, our model achieves $53.9\%$ on $\operatorname{mAP}_{role}$, outperforms the previous works. Considering the relatively small scale of the V-COCO dataset may impair the representation capability of the trained semantic embeddings, we test the results using the matching strategy without the semantic embeddings.

The Table~\ref{tb:hoia} also illustrates our effectiveness on the HOI-A test set. We reach a mAP of $72.19\%$, better than all the previous methods, including the method which adopts a relatively heavy-weight Hourglass-104 as backbone.
\vspace{-1.5mm}\subsection{Ablation Study}
\vspace{-0.5mm}
\noindent\textbf{Matching Strategy.}
Two variants of inference matching methods are implemented. As shown in Table~\ref{tab:ablation:match}, when only using the vector matching distance $\mathcal{D}$, or only using the semantic embedding distance $\mathcal{R}$ in Section~\ref{sec:training and postprocess}, the effectiveness are both compromised.

\vspace{0.5mm}\noindent\textbf{Semantic Embedding Settings.}
To explore the suitable semantic embedding setting, we evaluate the models with different embedding dimension $K$ and weight coefficient $\lambda_{\operatorname{emb}}$ of the training losses $\mathcal{L}_{\operatorname{pull}}$ and  $\mathcal{L}_{\operatorname{push}}$. As shown in Table~\ref{tab:ablation:K}, the effectiveness of our model is not sensitive to the embedding dimension. As $K$ changes from $4$ to $32$, the changing of the mAP result is only $0.66$ point. The embedding dimension $K$ is set to $8$ regarding the trade-off for both effectiveness and computational cost. As illustrated in Table~\ref{tab:weight}, the model performs best when training with $\lambda_{\operatorname{emb}}=0.1$, while the effectiveness will be impaired when $\lambda_{\operatorname{emb}}$ is increased or decreased.

\vspace{0.5mm}\noindent\textbf{Single Branch Variant.}
We implemented a single branch variant to detect instances along their interactions while keeping all hyper-parameters. As shown in Table~\ref{tab:ablation:component}, the variant achieves $25.91\%$ mAP on the HICO-DET dataset, which is $2.96\%$ lower than our AS-Net. Especially, the Rare mAP is $17.88\%$, which is $6.37\%$ lower than ours.  We consider it is because detection and interaction rely on some \textit{different features}. Lacking the interaction-related features such as human postures, the single branch variant is more likely to infer actions that frequently appear in the presence of the detected objects.

\vspace{0.5mm}\noindent\textbf{Basic Model.}
To verify the effectiveness of the basic framework, we implement a variant consists of one $6$-layer instance detection branch and one $6$-layer interaction detection branch, without the instance-aware interaction attention module and semantic embedding. Table~\ref{tab:ablation:component} articulates that our basic model~(Basic Model, Int$\times6$) achieves $27.52\%$ mAP on the HICO-DET dataset, which outperforms the previous methods by a large margin. 

\vspace{0.5mm}\noindent\textbf{Instance-aware Attention.}
Two other variants are evaluated by utilizing the instance-aware attention module to verify the contribution of branch aggregation. As presented in Table~\ref{tab:ablation:component}, the instance-aware attention module on our basic model~(+ IA Attn$\times6$, Int w/o emb$\times6$) improves mAP by $0.44$ point. For the basic model with the semantic embeddings~(+ Int w/ emb$\times6$), the improvements are $1.12$ points using the instance-aware attention module~(+ IA Attn$\times6$, Int w/ emb$\times6$). Therefore, we conclude that the instance-aware attention features from the instance branch are valuable for the interaction prediction. 

\vspace{0.5mm}\noindent\textbf{Semantic Embedding \& Instance-aware Attention.}
The basic model with the semantic embeddings~(+ Int$\times6$ w/ emb) improves slightly comparing to the basic model~(Basic Model, Int$\times6$) without the instance-aware attention as shown in Table~\ref{tab:ablation:component}. As the bridge connecting predicted instances and interaction vectors, the semantic embeddings also contribute to the training. However, the semantic embedding is less powerful than the instance-aware interaction attention module from the results. Based on the basic model with the semantic embedding, several variants are implemented, which consist of different interaction decoder layers or attention modules additionally:1)~$3$ instance-aware attention modules with $6$-layer interaction decoder~(+ IA Attn$\times3$, Int w/ emb$\times6$), performs attention every other layer; 2)~$3$ instance-aware attention modules with $3$-layer interaction decoder~(+ IA Attn$\times3$, Int w/ emb$\times3$). From Table~\ref{tab:ablation:component}, the performance is improved by about $1$ point utilizing the attention module and the semantic embedding jointly. Besides, it's better to use the instance-aware modules and the decoder layers with the same number of times. The effectiveness reduces slightly when we utilize the two modules with less times, while the amount of the model parameters is reduced significantly.
\vspace{-1.5mm}\subsection{Qualitative Results}
\begin{figure}[ht]
  \centering
  \includegraphics[width=1.0\linewidth]{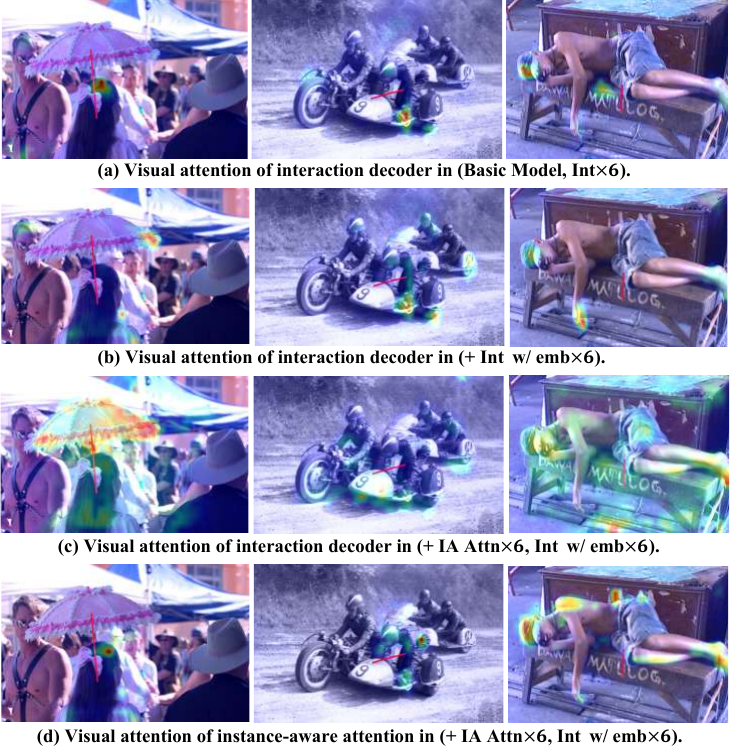}
  \vspace{-4.5mm}
  \caption{Visualization of the interaction-relevant attention. In each sub-figure, the interaction vector in red is pointing from the corresponding human center to the object center.
  }\vspace{-1mm}
  
  \label{visualization}
\end{figure}

\vspace{-1mm}
As shown in the first three rows of the Figure~\ref{visualization}, we visualize the interaction decoder attention for some interaction pairs in our basic model, the basic model with the semantic embeddings~(+ Int w/ emb$\times6$) and the model with both the instance-aware attention module as well as the semantic embeddings~(+ IA Attn$\times6$, Int w/ emb$\times6$), respectively. We also visualize the instance-aware attention in the last row for each example interaction pairs to present how the attention module contributes to the interaction prediction. 

From the Figure~\ref{visualization}~(a), the basic model without any branch aggregation focuses on some scattered redundant feature regions and leave out some interaction-relevant features. From the Figure~\ref{visualization}~(b), the model with semantic embeddings only partially alleviates the problem. For example, there is a girl holding an umbrella in the figures in the first column. To predict such interaction, the basic model concentrates on the head of the girl and the body of an irrelevant person. Correspondingly, the model with semantic embeddings pays attention to the edge of the umbrella and the body of the girl, while still concentrates on an irrelevant person. When we involve the instance-aware attention module, as shown in Figure~\ref{visualization}~(c) and Figure~\ref{visualization}~(d), the interaction branch concentrates on the whole umbrella and some body parts which are close to the umbrella, and the instance-aware attention module focuses on the body and head of the girl. In such a separated focus mechanism, our model can concentrate on the features more accurately.
\vspace{-3.5mm}\section{Conclusion and Future Work}
\vspace{-1.5mm}
In this paper, we reformulate HOI detection as an adaptive set prediction problem and propose a novel one-stage HOI detection framework, namely AS-Net. By aggregating interaction-relevant features from global contexts, and matching each ground-truth with the interaction prediction, our method demonstrates adaptive ability on both feature aggregation and supervision. Moreover, the designed instance-aware attention module contributes to intensify the instructive instance features, and we also introduce semantic embeddings to improve performance. The ablation studies verify the effectiveness of each key component of our model. Our AS-Net outperforms all existing methods on three HOI detection datasets. In the future, we plan to extend AS-Net to handle more general association problems, \eg, visual relationship detection and multi-object tracking.

\vspace{-4mm}
\paragraph{Acknowledgement} This work was partially supported by Sensetime Ltd. Group, National Natural Science Foundation of China under Grant 61876177, Zhejiang Lab (No. 2019KD0AB04),  Beijing Natural Science Foundation 4202034 and  Fundamental Research Funds for the Central Universities.
{\small
\bibliographystyle{ieee_fullname}
\bibliography{egbib}
}

\end{document}


\title{Supplementary Materials: Reformulating HOI Detection \\ as Adaptive Set Prediction}

\maketitle
\section{Visualization of Results}
To verify the effectiveness of our AS-Net, we select some examples from three HOI detection datasets HICO-DET~\cite{chao2018learning}, V-COCO~\cite{gupta2015visual} and HOI-A~\cite{liao2020ppdm}. From Figure \ref{fig:result}, when the human instance is far away from the corresponding object instance~(column 1), interactions locate close to each other~(column 2, 3), occlusion exists among instances~(column 4) or the ground-truth interaction categories are comparatively rare~(column 5), our model can still predict HOI interaction triplets accurately for each figure.
\begin{figure*}[ht]
  \includegraphics[width=1.0\linewidth]{figures/result_compressed.pdf}
  \caption{Visualization of some examples of our HOI prediction results. For each figure, the top three predicted HOI triplets are preserved. Labels of the human instances are in red, object labels are in green. Top three interaction categories in one figure are in blue~(\#1), yellow~(\#2) and purple~(\#3) respectively. Each predicted interaction vector is in blue, points from the human center to the corresponding object center.}
  \label{fig:result}
\end{figure*}

\section{Source Code}
We provide our implementation code based on PyTorch in the folder \emph{ASNet-eval.zip}. After the review period, we will release the full code with training codes, data and checkpoints on GitHub. The link of our released code will be included in the final version.

{\small
\bibliographystyle{ieee_fullname}
\bibliography{egbib}
}